\documentclass{article}

\usepackage{arxiv}

\usepackage[utf8]{inputenc} 
\usepackage[T1]{fontenc}    
\usepackage{hyperref}       
\usepackage{url}            
\usepackage{booktabs}       
\usepackage{amsfonts}       
\usepackage{nicefrac}       
\usepackage{microtype}      
\usepackage{cleveref}       
\usepackage{lipsum}         
\usepackage{graphicx}
\usepackage{natbib}
\usepackage{doi}

\title{Can the language of the collation be translated into the language of the stemma? Using Machine Translation for witness localization}

\author{Armin Hoenen \\
	Department of Empirical Linguistics\\
	Goethe University Frankfurt\\
	Frankfurt, Germany \\
	\texttt{hoenenarmin@gmail.com} } 

\renewcommand{\shorttitle}{\textit{arXiv} Template}

\hypersetup{
pdftitle={Hoenen2022Canthelanguage},
pdfsubject={stemmatology},
pdfauthor={Armin Hoenen},
pdfkeywords={stemmatology, phylogeny},
}

\begin{document}
\maketitle

\begin{abstract}
Stemmatology is a subfield of philology where one approach to understand the copy-history of textual variants of a text (witnesses of a tradition) is to generate an evolutionary tree. Computational methods are partly shared between the sisterdiscipline of phylogenetics and stemmatology. In 2022, a surveypaper in nature communications\footnote{\url{https://www.nature.com/ncomms/}} found that Deep Learning (DL), which otherwise has brought about major improvements in many fields \cite{krohn2020deep} has had only minor successes in phylogenetics and that ''it is difﬁcult to conceive of an end-to-end DL model to directly estimate phylogenetic trees from raw data in the near future'' \cite[p.8]{sapoval2022current}. In stemmatology, there is to date no known DL approach at all. In this paper, we present a new DL approach to placement of manuscripts on a stemma and demonstrate its potential. This could be extended to phylogenetics where the universal code of DNA might be an even better prerequisite for the method using sequence to sequence based neural networks in order to retrieve tree distances.
\end{abstract}


\section{Introduction}
The disciplines of philology, biology and linguistics are amongst those which have a branch occupied with the reconstruction of genealogical trees retrieving and showing the evolutionary relationships between certain of their domainspecific units (DNA of beings, texts of manuscripts, languages), see also \cite{roelli2020handbook}. The subdisciplines stemmatology (philology), phylogenetics (biology) and historical linguistics developed and partly shared methodology. Whilst in phylogenetics and linguistics, see \cite{suvorov2020accurate,wu2020automatically}, Deep Learning (DL) has been applied in certain studies, the same cannot be said for stemmatology. This might be unfortunate looking at the successes of DL in other disciplines, such as machine vision \cite{krohn2020deep}. Hoewever, a recent survey paper on the application of DL in the biosciences found, that DL was least successful in phylogenetics, that some of the few proposed methods were not able to reach the performance of exisiting ones \cite{zaharias2022re} and that 'it is difﬁcult to conceive of an end-to-end DL model to directly estimate phylogenetic trees from raw data in the near future'' \cite[p.8]{sapoval2022current}. For stemmatology, some specific features of the units of interest (mainly manuscripts and the texts they hold) constitute a difference to each of its sister sciences, which is why developing a DL approach here could still be worthwhile. In this paper, after a brief recapitulation of related literature, DL approaches for stemmatology will be discussed in general before presenting an experiment where DL machine translation technology is being used for phylogenetic placement which is evaluated on a so-called artificial dataset, that is a dataset where a text has been manually copied and recopied and then aligned and digitized later on \cite{Spencer:Davidson:Barbrook:Howe:2004}. 

\section{General Discussion on DL in stemmatology}
\label{sec:gd}
\subsection{DL in adjacent disciplines}
Whilst to the best knowledge of the author, so far DL has not been applied to stemmatology, the closely related sister-disciplines of phylogenetics (biology) and historical linguistics feature some research using DL. Between these and other disciplines using domain-specific input data for the generation of genealogical trees, there are various commonalities, but also various differences which sometimes prevent a direct transferability. For instance do biology and linguistics feature an assumed tree of all life resp. of all languages if one assumes one single origin of life and language. This cannot be said for manuscripts or texts. There is no such thing as a primeval text from which all other texts derive. On the other hand, lateral gene transfer, which is the transfer of single genes even across species boundaries, is similar to so-called contamination which is the transfer of certain parts of a text from another than the principle model, since for instance there is a page missing in the model. For more details on these relations with other fields consider (\cite{roelli2020handbook}). Thus, in order to find DL approaches which could be applicable to stemmatology, it could be worthwhile to survey adjacent disciplines.

In linguistics, DL has been applied using information from multilingual or cross-lingual ressources, see for instance \cite{Eger:Hoenen:Mehler:2016}. \cite{rama2020probing} use such DL models to extract genetic relationships between languages. However, this approach is not transferable since there the units of genealogical information in stemmatology are rather variations which are only definable as differences between at least two texts and not single words. Consequently, a multivariational model similar to a BERT model \cite{kenton2019bert} is not constructable for stemmatology, apart from the difference in abundance of language data and stemmatological data. 

In phylogenetics on the other hand, recently, as mentioned above, a survey paper from nature \cite{sapoval2022current} has investigated DL approaches across the biosciences and found few approaches in phylogenetics. The reason is in part that an ideal input-output schema for DL in stemmatology and phylogenetics likewise would be the multiple sequence alignment (or in stemmatology collation) and the output the corresponding tree topology. Now, as \cite{Felsenstein:1978} has shown for bifurcating and multifurcating trees and \cite{HEG:2017} elaborated and connected to stemmatology for Greg trees \cite{Flight:1990} - a mathematical model for stemmata - the number of possible topologies grows rapidly into unfeasible numbers. For 10 manuscripts, there would already be 102 515 201 984 possible trees. Such large numbers are unusable as output dimensions/classes for neural networks, computationally and in terms of accuracies (think of the size of the confusion matrices and the probabilities of a true hit, the required numbers of training instances etc).

What is possible however is to use very low dimensions, such as trees with 4 species, where there exist only 4 undirected tree topologies. \cite{suvorov2020accurate} and \cite{zou2020deep} used such an approach. A problem was to have adequate data. Authors used simulated datasets since \cite{sapoval2022current} (relating to \cite{nute2019evaluating}) found that
\begin{quote}
classifiers like DL models require training data, and benchmark data where the true phylogeny is known is
almost impossible to obtain in this field [phylogenetics]. Instead, simulations have been the method of choice for generating training data, but this is a major dependency and methods are known to have divergent performance on simulated and biological data.
\end{quote}
Consequently, for similar approaches to stemmatology, one would need a device simulating a lot of data, since the availability of such data for evaluation is also limited in stemmatology \cite{roelli2020handbook} in the first place. Apart from this, it is not clear if such a trained model would be very useful in stemmatology, since in contrast to biology where there is a universal code for DNA, the words constituting texts are of a completely different nature.

Yet other disciplines with major similarities to stemmatology are largely smaller, dependent on stemmatology or less technical than is stemmatology, \cite{roelli2020handbook}. Thus, although there are what has been called 'minor successes' \cite{sapoval2022current} of DeepLearning in adjacent disciplines, those seem either hardly transferable or not potent enough to start adapting them for stemmatological needs.

\subsection{Reinforcement learning - the philologists game}
Another very successful application involving DL was in Google DeepMind's AlphaGoZero. This technology now allows the algorithm to outperform humans in board games of a large complexity, such as Go. If stemmabuilding could be defined as if it were a game, maybe such a technology could be used to solve the problem of generating a genealogical tree automatically from a collation. Certain problems in information theory are equivalent to each other requiring a solution to one of them and a reformulation. 

There could be various ways to reformulate stemmatology as a game. Either a self-game (case 1), then AlphaGoZero would not be directly applicable, or a competitive game with a non-shared field (the stemma) (case 2), that is two or more players build their own stemma and get a final score each or finally, a shared field - that is all build the stemma together one step at a time, but in the end the score would have to be divided corresponding to the quality of the contributions (case 3). To exemplify one of the possibilities (case 1), we spell ist out
\begin{enumerate}
\item start from a random edge you draw between any two witnesses with recurrence to their texts (first move)
\item successively attach more nodes never violating a DAG (move rules) until no more nodes left
\item receive a score equal to the TreeEditDistance to the true stemma (there are far fewer TEDs than tree topologies)
\end{enumerate}
 
A principal benefit of this thought experiment may be to understand that there are much fewer possible TeDs than there are possbile topologies and that it may not matter as much which wrong topology one produces than how wrong it would be overall. Albeit from a TeD, we cannot reconstruct a tree.

However, each of the cases would require some non-trivial adaptation of the technical backend of AlphaGoZero with an uncertain perspective of realizability. More importantly, once one would have such a model trained, it would only be applicable to the same input data, that is, the same text. This is because variation has to be regarded with context, meaning local context (the word before and after) and global context (the other variants other manuscripts which carry the same text display at that position). Thus, the distance implied by a variant is always dependent on global and local context of the current text and it is improbable that one could train an algorithm of this kind to apply it then to input data other than yet another variant of the training text. This means such an endeavour could not solve the problem of building a complete tree from scratch for otherwise new unseen traditions.  
 
\subsection{Approaches which could be transferred}
In phylogenetics, \cite{suvorov2020accurate,zou2020deep} presented an approach on artificial data for 4 taxa trees. In principle this could be transferred to stemmatology if subsampling all possible 4 witness subtrees of a larger stemma.
That would again limit the applicability to texts of the same tradition but it would be applicable at least to solve the placement of new manuscripts (see below). However, \cite{sapoval2022current} concerning the approach state, that:
\begin{quote}
Recently CNNs have been used to infer the unrooted phylogenetic tree on four taxa [...] an analysis of the
performance of the method [...] shows that CNNs were not as accurate as other standard tree estimation methods,
e.g., maximum likelihood, maximum parsimony, and neighbor joining.
\end{quote}
Against this background, it seems hardly promising to transfer such an approach to stemmatology where maximum likelihood, maximum parsimony and neighbor joining have been used continuously. However, in the appendix, a small prototype of a generator for simulated stemmatological data is being outlined which could for future experimentation generate large numbers of artificial data. A first example corpus has been created and is described in the appendix. 

On the other hand there is another approach to a subproblem applying to both stemmatology and phylogenetics: the placement of new sequences or in biological terms: \emph{phylogenetic placement}. Whilst in biology, this is most often needed to localize newly found species amongst the family, order etc. of organisms they belong to or broadlier speaking on the tree of life, presumably in stemmatology it could be used to verify the assumed place of a witness on a stemma. To this end, \cite{howe2001manuscript} have used placement, especially as a demonstrator to identify successive contamination, where one part of the manuscript has been copied from one model (vorlage) and another from another, so that the different parts cluster differently.

Lateral gene transfer for instance through microbial vectors is a phenomenon quite similar to contamination leading to parts of a genome having a slightly different provenience and genealogy than the rest. 
A DL solution to the problem of gene vs. species tree has been proposed by \cite{jiang2021depp} and seems to be a more successful application of DL in phylogenetics. This approach could be transferable to stemmatology, yet the way in which it solves the gene vs. species tree problem or rather the result may have no real precedents in stemmatology and would thus have to be critically reflected upon by the philologists community. It optimizes the resulting tree to represent the species relations more genuinely modifying relations also by information from gene relations. If in stemmatology, a tree of this kind is preferred or if - as current practice indicates - rather two or more trees for each text segment are preferred is not entirely clear. In this paper, we use another DL approach for phylogenetic placement.

\section{Main Experiment}
\label{sec:ex}

\emph{Phylogenetic placement} is the task of localizing a sequence on an existing tree within the genealogical relations of similar items, see for instance \cite{mirarab2012sepp}. For stemmatology analogously 'witness placement' is the task of placing a new or uncertain textual witness amongst the other witnesses. Since in philology philological judgement (operationalized as \emph{iudicium} in stemma building, \cite{roelli2020handbook}) for instance about the weights/meaning of certain variants is highly esteemed, on the other hand some patterns of variation naturally seem more conclusive than others. It so can happen that for the philologist compiling a stemma for the preface of a possibly critical edition some manuscripts exact localization on the stemma is unsure/less sure than others. In this case phylogenetic placement may serve as a test and corroborate or question the philological decision. The case of new witnesses appearing also happens. In that case however both the alignment and the placement must be achieved, whereas in the former case a complete alignment containing the insecure node is preexisting. Placement alone would not solve alignment, but aligment alone is the prequisite for automatic tree generation (manually tasks can be intertwined). In this paper, we present a DL approach to placement and evaluate it on leaf nodes of a manually produced tradition for which the true stemma is known. This scenario starts from a complete collation.

\section{Related Work}
\label{sec:rw}
\cite[p.247/248]{mirarab2012sepp} define phylogenetic placement with respect to previous research formally as follows

\begin{itemize}
\item Input: the backbone tree $T$ and alignment $A$ on set $S$ of full-length sequences, and query sequence $s$.
\item Output: tree $T^{\prime}$ containing $s$ obtained by adding $s$ as a leaf to $T$.
\end{itemize}

and summarize methods which had until 2012 been used satisfying the following two steps
\begin{enumerate}
\item Step 1: insert $s$ into alignment $A$ to produce the extended alignment $A^{\prime}$.
\item Step 2: add $s$ into $T$ using $A^{\prime}$, optimizing some criterion.
\end{enumerate}

The methods and softwares they describe did not use DL methods, neither a more recent approach by \cite{balaban2020apples} which allows placement based on distances even without alignments. \cite{sapoval2022current} identified only \cite{jiang2021depp} albeit with some criticism to the mathematical fit of the method to reality.

Finally \cite{howe2001manuscript} use tree generation for different portions of a sequence alignment to show that the place of a particular manuscript is different depending on the portion reflecting successive contamination (shift of exemplar). Note however that in their approach, there is no backbone tree, that is the trees of the different portions do differ not only with respect to the placement of the manuscript under scrutiny.
In contrast to that, we adhere to the above outlined definition but start from an already present extended alignment where we ``blind out'' or remove or hold back certain nodes from the known true tree, in order to then insert them by our approach and then evaluate localization.

\section{Methods \& Materials}
\label{sec:mm}

The experiment was conducted on a single personal computer with a GPU. It uses a technology which is normally used for a quite different field: machine translation, where however also one sequence is turned into another.

\subsection{Approach}
We use a \textbf{sequence to sequence} DL approach, the steps of which are as follows:

\begin{itemize}
\item from collation and corresponding true tree, hold back any one leaf node
\item train a DL model with input $=$ pairwise variant configuration, output $=$ distance (number of edges on stemma)
\item let model estimate length between held back node and any other on the tree
\item use a scoring to place node from estimates
\begin{itemize}
\item only one node truly has edge dist 1: the parent; if also only one in the DL estimates has this value, take that node
\item else: for each node $+$ estimate localize and score each node which has the predicted distance and could thus be the predicted parent; retrieve score winner over all predictions
\end{itemize}
\item evaluate placement and estimate accuracy
\end{itemize}

\begin{figure}
\centering
  \includegraphics[scale=0.5]{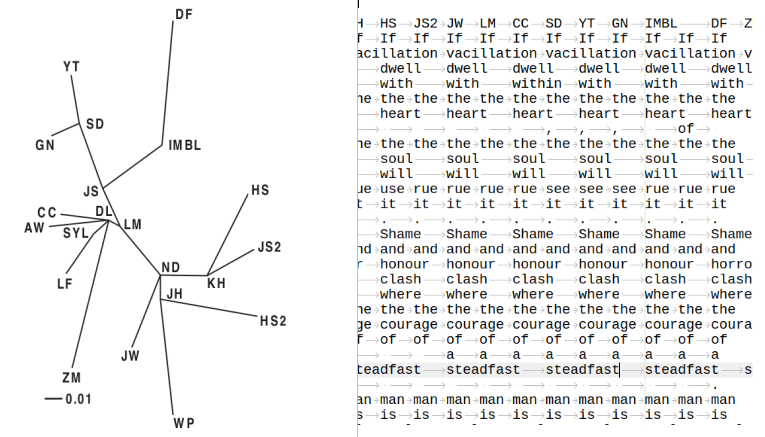}\\
  \includegraphics[scale=0.5]{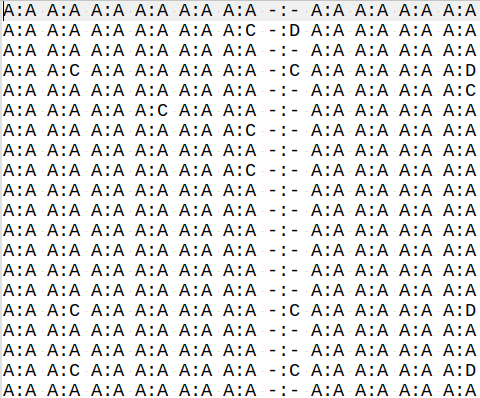}
  \caption{The true stemma of the Parzival tradition alongside the collation is being transformed into the input for the machine translation DL algorithm. The first line displays the comparison of the first two columns in the collation. Each letter:letter `word' is one row from the collation, so where the input reads A:A both manuscripts have the same original variant.}
  \label{fig:in}
\end{figure}

The exact input was extracted from the collation (multiple sequence alignment). Figure \ref{fig:in} shows the transformation. The output was directly taken from the known stemma and is just one digit.

\subsection{Dataset}
For evaluation, we use the Parzival dataset \cite{Spencer:Davidson:Barbrook:Howe:2004}, features of which are given in table \ref{tab:table1}. There are some more such traditions \cite{Baret:Mace:Robinson:2004,Roos:Heikkila:2009,Hoenen:2015a} but many of them have either cycles or consist of various independent trees, which makes Parzival the most applicable dataset of this kind.

\begin{table}
	\caption{Features of the Parzival tradition}
	\centering
  \begin{tabular}{ l | c | l }
    \textbf{Datatype}&\textbf{Value}&\textbf{Note} \\
    \hline
    \emph{Language }& English (from German, archaic) &  \\ 
    \emph{Copy-Mode} & manual, written & no contamination \\ 
    \emph{n. of mss} & 21 &  \\
    \emph{n. of leafs} & 12 &  \\
    \emph{n. of rows} & 958 &  \\
    \emph{max dist leaf-leaf} & 6 & edges \\
    \emph{Publication} & \cite{Spencer:Davidson:Barbrook:Howe:2004} &  \\
    \hline
  \end{tabular}
	\label{tab:table1}
\end{table}

\subsection{Computation}

As for the computation, it was run on a computer with an i9 Intel processor and a 4 GB NVidia GForce GPU as well as sufficient RAM (32GB). Pre- and postprocessing of the data was done by some Java code as was evaluation, whilst training and estimation was done with pythons ONMT framework \cite{klein2018opennmt} in a pytorch implementation. 
Since in the training of neural networks GPU time and memory can be somewhat of a limiting factor, not all parameter combinations could be tested. One run for the Parzival tradition required roughly 10 hours GPU time. Testing more neural network architectures and parameterizations remains for future research. 

\subsection{Pipeline}
First, the data from Parzival, which was already present in the form of an alignment/collation had to be transformed and partitioned for each run. Here, for each possible witness pair a phrase, that is one line of input was generated in such a way that 
\begin{itemize}
\item binary input was used: true if both variants are the same, false otherwise
\item variant letters, see Figure \ref{fig:in}
\item actual words of the collation (the:the magpie:magpie ...) 
\end{itemize}
were used.

Then data was partitioned into evaluation data: for each leaf, all edges involving this node were held back and training data: from the remaining witness pairs, 5 or 10 were held back as validation data. Since with 190 node pairs (paths) per held back node left, there was not much trainingdata in terms of instances, the validation set size was chosen respectively small.

Then the training set was forwarded to OMT to be trained. At first a default parameterization was chosen and encoded into a config yaml file. For each parameter variation, another model was trained by other yaml files. 

After training was completed, we used the models to ``translate'' all 20 pairwise inputs involving the held-back node for all 12 held-back nodes (summing up to 240 estimates). 
We then evaluated the estimates and computed for each node the placement position which again was compared to the original one and its distance on the true tree.

Additionally, we generated a random baseline by producing 1000 times random estimate runs in the same range as the output vocabulary. 

\section{Results}
\label{sec:res}

Due to various constraints on possible GPU times, we did not do a full grid search through all possible parameterizations. We tested however parametrizations and input configurations from very different ends of the possibility space so to speak and rather concentrated on interpreting the results more profoundly. 

\subsection{Architectures \& Input Configurations}
We tested three neural network architectures: Bi-LSTMs, BRNNs and Transformers but found that for our current explorative scenario only BRNNs were quick and accurate enough to allow for the exploration of different settings. The other two required far more GPU time with flatter accuracy curves when parameterized roughly comparably. It might be that future research can show that one of the other architectures produces (slightly) more accurate results.

Various input variations were tested and compared

\begin{itemize}
\item \textbf{INPUT-DIFF-TYPE}: binary input, variant letters (sorted and unsorted),\footnote{The sorted `incarnation' of C:A would be A:C.} actual words [not tested: letter-correspondences]: \textbf{variants sorted} performed best
\item \textbf{INPUT-TYPE}: only places of variation vs. all places: \textbf{all places has an advantage}
\item \textbf{INPUT-SETSIZE}: validation set size: 5-185 and 10-180: \textbf{no significant difference}
\item \textbf{ARCHITECTURE}: rnn and wordvec sizes: \emph{slight hints that lower dimensions may work better in this setting but could distinguish less between possible placements (depending on tradition size probably)}, number of layers: more than 1 layers significantly slowed down training while at the same time the validation accuracy curve did for a test case not suggest substantial improvement.
\end{itemize}

The number of training steps was determined as the best accuracy curve cutoff as per validation accuracy across all nodes.

Maintaining some default values, we decided for the following parameterization as our principle testing architecture for which we made the comparisons of input types etc.: BRNN with 512 rnn-size, 128 word vec size, 1 layer, no dropout, 5:185 validationset proportion, batch-size 16, 7000 training steps. It produced the following result:

\begin{center}
  \begin{tabular}{ l | c | l }
    \textbf{Feature}&\textbf{Value}&\textbf{Note} \\
    \hline
    \emph{correct predictions}& 111/240 ($0.46$) & best $135 (0.56)$ \\ 
    \emph{average deviation} & $0.6$ (SD:$0.6$) & best $0.5$ (SD:$0.6$) \\ 
    \emph{max dist} & 3 & best 2\\
    \emph{hitrate localizations} & $9.5/12$ ($\mathbf{0.79}$) & distance of 2 misclassified  \\
     &  & from parent: 1  \\
    \emph{average radius} & $0,23$ & around true parent\\
    \hline
    \textbf{Random baseline} & \textbf{100.000 iterations} & averages\\
    \hline
    \emph{correct predictions}& 40/240 ($0.17$) & max 67, diff from above significant \\ 
    \emph{average deviation} & $1.85$ (SD:$1.4$) &  \\ 
    \emph{max dist} & $5$ & 5 is absolute possible maximum, 10 times not 5 \\
  \end{tabular}
\end{center}

Placement of 9 leaf nodes was perfect, one had two equiprobable candidates whereof one was the original parent, the other a sibling and the 2 mislocated leafs were only 1 edge away from their original parent. For the random baseline, there was no reason to compute the corresponding values since already the estimation values show that this would not lead to feasible results.


\section{Discussion}
\label{sec:dis}

Since edge distances are natural numbers and in our scenario even strings, the network is forced to decide for one output even though the activation for the second strongest could be almost indistinguishable. This indicates that if the output were a discrete number, the estimates could show an even better fit. If a large proportion of the cases where the distance estimate is only 1 off would be resolved by using a number instead of a string, the percentage of correct estimates could even rise significantly. This could easily be achieved by replacing the output layer by a single neuron through a softmax. There are cases, where the MT overgenerated and produced repeated sequences. These cases were rare (up to 4 of 240 predictions were observed such; for evaluation we then took the first) but would also not occur in a one-neuron output. The problem that an edge dist is only expressable as a natural number would however remain the same. 

However, despite a relatively low ratio of true estimation hits, the localization algorithm still places the leafs correctly in our example 80\% of the time and with a radius from the true parent of $0.23$ or in other words, if the algorithm errs, it is off only slightly, usually by 1 edge. This dataset was relatively small and for larger stemmata there is also more training data, a deterioration is thus not to be expected. As furthermore, the comparison with a random baseline could show, the method works.

This entails that in the collation there is edge-distance information, not just text distance information. Thus, the information from a pairwise comparison of sequences not only constitutes a textual distance, but in the context of the complete collation carries information of the distance on the true tree. The DL method we presented is a way to extract such information from the data. 
The output in our method is however encoded as a string, not a value, but since the strings exclusively represent this value such a data transformation works well. 
We found that an input which encodes all-places of the collation for pairwise comparisons is producing better results than if we only use places-of-variation. This points to some of the information being encoded within what and how much is shared (which a computer might exploit with more ease and more objectively than humans who read for difference).

The most crucial question was however why the sorted encoded variants performed best and not for instance the words. After all, this is a simplification where the same letter A comes to stand for `original variant' no matter what the actual word at the corresponding position of the collation is. One can now ask oneself why such a generalization might work? One explanation could be to postulate something like a `concentration profile' of scribes which is most informative. This would imply that it is not the exact `error' or deviation that a scribe introduces when at a certain position of the collation. There is plenty of possible variation to introduce (all the more if one include non-word errors and the like) at any point in text time. It might matter more where a place of variation is found. Each copy process could then have its own profile. Corrections on the one hand which can happen and reintroduce original variants and slips of the pen on the other are both part of this concentration profile. Now, both are distinguishable by direction. This direction of deviation would not be taken into account if sorting variant letters. That could be a reason why this input configuration worked best. If this implies generalizability across traditions is an interesting follow-up question about which we conducted a small follow-up experiment to see if such a presupposition could hold. 
We used the Parzival AW-7000 model for the much smaller NB (French, 13 nodes, 6 leafs) tradition and got a result worse than the comparative random baseline but with at least lower maximum distance. Then, we used the same model of the same node for classifying a subtree of the larger Heinrichi tradition (Finnish, 67 nodes, 35 leafs) and again got no usable result. Finally, we used a Heinrichi node model to classify all Parzival nodes with a slightly but significantly better than chance output. Whilst these results are hard to interpret, they show a prospect for future research where DL models trained on a large enough tradition could lead to transferability across traditions. This in turn could be a first step towards possible tree generation through DL. 

\section{Conclusion}
\label{sec:con}
We have discussed some of the problems DL approaches suffer in stemmatology and related disciplines. We then demonstrated a DL approach for stemmatology based on machine translation, albeit for phylogenetic placement not tree generation, with at least moderate success. The approach is not yet entirely optimized. However, it correctly placed the vast majority of leaf nodes in the Parzival dataset. 
We found that a collation must carry information not only about pairwise distances but also about edge distances on the true stemma. 
We discussed various interpretations of our results and questions the approach implies for future research, such as if there is transferability of the extracted information to other traditions.

\bibliographystyle{natbib}
\bibliography{Hoenen2022CanTheLanguage.bib} 

\begin{thebibliography}{27}
\providecommand{\natexlab}[1]{#1}
\providecommand{\url}[1]{\texttt{#1}}
\expandafter\ifx\csname urlstyle\endcsname\relax
  \providecommand{\doi}[1]{doi: #1}\else
  \providecommand{\doi}{doi: \begingroup \urlstyle{rm}\Url}\fi

\bibitem[Balaban et~al.(2020)Balaban, Sarmashghi, and
  Mirarab]{balaban2020apples}
Metin Balaban, Shahab Sarmashghi, and Siavash Mirarab.
\newblock Apples: scalable distance-based phylogenetic placement with or
  without alignments.
\newblock \emph{Systematic Biology}, 69\penalty0 (3):\penalty0 566--578, 2020.

\bibitem[Baret et~al.(2004)Baret, Mac\'{e}, and
  Robinson]{Baret:Mace:Robinson:2004}
P.~Baret, C.~Mac\'{e}, and P.~Robinson.
\newblock Testing methods on an artificially created textual tradition.
\newblock In \emph{Linguistica Computationale XXIV-XXV}, volume XXIV-XXV, pages
  255--281, Pisa-Roma, 2004. Instituti Editoriali e Poligrafici Internationali.

\bibitem[Damerau(1964)]{Damerau:1964}
Fred~J. Damerau.
\newblock A technique for computer detection and correction of spelling errors.
\newblock \emph{Communications of the ACM}, 7:\penalty0 171--176, 1964.

\bibitem[Eger et~al.(2016)Eger, Hoenen, and Mehler]{Eger:Hoenen:Mehler:2016}
Steffen Eger, Armin Hoenen, and Alexander Mehler.
\newblock Language classification from bilingual word embedding graphs.
\newblock In \emph{Proceedings of COLING 2016}. ACL, 2016.

\bibitem[Felsenstein(1978)]{Felsenstein:1978}
Joseph Felsenstein.
\newblock The number of evolutionary trees.
\newblock \emph{Systematic Zoology}, 27\penalty0 (1):\penalty0 27--33, 1978.

\bibitem[Flight(1990)]{Flight:1990}
Colin Flight.
\newblock How many stemmata?
\newblock \emph{Manuscripta}, 34\penalty0 (2):\penalty0 122--128, 1990.

\bibitem[Geyer(1977)]{Geyer:1977}
LH~Geyer.
\newblock Recognition and confusion of the lowercase alphabet.
\newblock \emph{Perception \& Psychophysics}, 22\penalty0 (5):\penalty0
  487--490, 1977.

\bibitem[Hoenen(2015{\natexlab{a}})]{Hoenen:2015a}
Armin Hoenen.
\newblock {Das artifizielle Manuskriptkorpus TASCFE}.
\newblock In \emph{{DHd 2015 - Von Daten zu Erkenntnissen - Book of
  abstracts}}. DHd, 2015{\natexlab{a}}.
\newblock URL \url{http://gams.uni-graz.at/o:dhd2015.abstracts-gesamt}.

\bibitem[Hoenen(2015{\natexlab{b}})]{Hoenen:2015nldb}
Armin Hoenen.
\newblock Simulating misreading.
\newblock In \emph{Proceedings of the 20th International Conference on
  Applications of Natural Language to Information Systems (NLDB)},
  2015{\natexlab{b}}.

\bibitem[Hoenen(2016)]{Hoenen:2016DH}
Armin Hoenen.
\newblock {Silva Portentosissima - Computer-Assisted Reflections on
  Bifurcativity in Stemmas}.
\newblock In \emph{Digital Humanities 2016: Conference Abstracts. Jagiellonian
  University \& Pedagogical University}, pages 557--560, 2016.
\newblock URL \url{http://dh2016.adho.org/abstracts/311}.

\bibitem[Hoenen et~al.(2017)Hoenen, Eger, and Gehrke]{HEG:2017}
Armin Hoenen, Steffen Eger, and Ralf Gehrke.
\newblock How many stemmata with root degree k?
\newblock In \emph{Proceedings of the 15th Meeting on the Mathematics of
  Language}, pages 11--21, 2017.

\bibitem[Howe et~al.(2001)Howe, Barbrook, Spencer, Robinson, Bordalejo, and
  Mooney]{howe2001manuscript}
Christopher~J Howe, Adrian~C Barbrook, Matthew Spencer, Peter Robinson, Barbara
  Bordalejo, and Linne~R Mooney.
\newblock Manuscript evolution.
\newblock \emph{TRENDS in Genetics}, 17\penalty0 (3):\penalty0 147--152, 2001.

\bibitem[Jiang et~al.(2021)Jiang, Balaban, Zhu, and Mirarab]{jiang2021depp}
Yueyu Jiang, Metin Balaban, Qiyun Zhu, and Siavash Mirarab.
\newblock Depp: deep learning enables extending species trees using single
  genes.
\newblock \emph{bioRxiv}, 2021.

\bibitem[Kenton and Toutanova(2019)]{kenton2019bert}
Jacob Devlin Ming-Wei~Chang Kenton and Lee~Kristina Toutanova.
\newblock Bert: Pre-training of deep bidirectional transformers for language
  understanding.
\newblock In \emph{Proceedings of NAACL-HLT}, pages 4171--4186, 2019.

\bibitem[Klein et~al.(2018)Klein, Kim, Deng, Nguyen, Senellart, and
  Rush]{klein2018opennmt}
Guillaume Klein, Yoon Kim, Yuntian Deng, Vincent Nguyen, Jean Senellart, and
  Alexander~M Rush.
\newblock Opennmt: Neural machine translation toolkit.
\newblock \emph{arXiv preprint arXiv:1805.11462}, 2018.

\bibitem[Krohn et~al.(2020)Krohn, Beyleveld, Bassens, and
  Lichtenberg]{krohn2020deep}
J.~Krohn, G.~Beyleveld, A.~Bassens, and K.~Lichtenberg.
\newblock \emph{Deep Learning illustriert: Eine anschauliche Einf{\"u}hrung in
  Machine Vision, Natural Language Processing und Bilderzeugung f{\"u}r
  Programmierer und Datenanalysten}.
\newblock dpunkt.verlag, 2020.
\newblock ISBN 9783960887522.
\newblock URL \url{https://books.google.de/books?id=3QT8DwAAQBAJ}.

\bibitem[Mirarab et~al.(2012)Mirarab, Nguyen, and Warnow]{mirarab2012sepp}
Siavash Mirarab, Nam Nguyen, and Tandy Warnow.
\newblock Sepp: Sat{\'e}-enabled phylogenetic placement.
\newblock In \emph{Biocomputing 2012}, pages 247--258. World Scientific, 2012.

\bibitem[Nute et~al.(2019)Nute, Saleh, and Warnow]{nute2019evaluating}
Michael Nute, Ehsan Saleh, and Tandy Warnow.
\newblock Evaluating statistical multiple sequence alignment in comparison to
  other alignment methods on protein data sets.
\newblock \emph{Systematic biology}, 68\penalty0 (3):\penalty0 396--411, 2019.

\bibitem[Rama et~al.(2020)Rama, Beinborn, and Eger]{rama2020probing}
Taraka Rama, Lisa Beinborn, and Steffen Eger.
\newblock Probing multilingual bert for genetic and typological signals.
\newblock \emph{arXiv preprint arXiv:2011.02070}, 2020.

\bibitem[Roelli(2020)]{roelli2020handbook}
Philipp Roelli.
\newblock \emph{Handbook of Stemmatology: History, Methodology, Digital
  Approaches}.
\newblock De Gruyter, 2020.

\bibitem[Roos and Heikkil\"{a}(2009)]{Roos:Heikkila:2009}
Teemu Roos and Tuomas Heikkil\"{a}.
\newblock Evaluating methods for computer-assisted stemmatology using
  artificial benchmark data sets.
\newblock \emph{Literary and Linguistic Computing}, 24:\penalty0 417--433,
  2009.

\bibitem[Sapoval et~al.(2022)Sapoval, Aghazadeh, Nute, Antunes, Balaji,
  Baraniuk, Barberan, Dannenfelser, Dun, Edrisi, et~al.]{sapoval2022current}
Nicolae Sapoval, Amirali Aghazadeh, Michael~G Nute, Dinler~A Antunes, Advait
  Balaji, Richard Baraniuk, CJ~Barberan, Ruth Dannenfelser, Chen Dun,
  Mohammadamin Edrisi, et~al.
\newblock Current progress and open challenges for applying deep learning
  across the biosciences.
\newblock \emph{Nature Communications}, 13\penalty0 (1):\penalty0 1--12, 2022.

\bibitem[Spencer et~al.(2004)Spencer, Davidson, Barbrook, and
  Howe]{Spencer:Davidson:Barbrook:Howe:2004}
Matthew Spencer, Elizabeth~A. Davidson, Adrian~C. Barbrook, and Christopher~J.
  Howe.
\newblock Phylogenetics of artificial manuscripts.
\newblock \emph{Journal of Theoretical Biology}, 227:\penalty0 503--511, 2004.

\bibitem[Suvorov et~al.(2020)Suvorov, Hochuli, and
  Schrider]{suvorov2020accurate}
Anton Suvorov, Joshua Hochuli, and Daniel~R Schrider.
\newblock Accurate inference of tree topologies from multiple sequence
  alignments using deep learning.
\newblock \emph{Systematic biology}, 69\penalty0 (2):\penalty0 221--233, 2020.

\bibitem[Wu et~al.(2020)Wu, Zhong, and Black]{wu2020automatically}
Peter Wu, Yifan Zhong, and Alan~W Black.
\newblock Automatically identifying language family from acoustic examples in
  low resource scenarios.
\newblock \emph{arXiv preprint arXiv:2012.00876}, 2020.

\bibitem[Zaharias et~al.(2022)Zaharias, Grosshauser, and
  Warnow]{zaharias2022re}
Paul Zaharias, Martin Grosshauser, and Tandy Warnow.
\newblock Re-evaluating deep neural networks for phylogeny estimation: The
  issue of taxon sampling.
\newblock \emph{Journal of Computational Biology}, 2022.

\bibitem[Zou et~al.(2020)Zou, Zhang, Guan, and Zhang]{zou2020deep}
Zhengting Zou, Hongjiu Zhang, Yuanfang Guan, and Jianzhi Zhang.
\newblock Deep residual neural networks resolve quartet molecular phylogenies.
\newblock \emph{Molecular Biology and Evolution}, 37\penalty0 (5):\penalty0
  1495--1507, 2020.

\end{thebibliography}

\begin{appendix}
\section{Simulating data}
\label{sec:ex}

Referring back to \cite{Hoenen:2016DH} we produced stemmata and then added a prototypical algorithm which, starting from one input text, copied these along the edges of the stemma. The copy process was not simulated as sophisticateldy yet a described in \cite{Hoenen:2015nldb} but with the following logic. Starting from root, each node gets an input text which gets copied by an artificial scribe confusing random letters according to a confusion matrix \cite{Geyer:1977} at a certain error ratio and mostly within class
vowel or consonants. If a word upon copying is not in a large english token lexicon (generated from subtitles word lists), a word with minimal Damerau-Levenshtein distance \cite{Damerau:1964} will be substituted as its correction (if more than one such words are in the lexicon, one is randomly chosen). 

For the Human Rights Declaration as input to a stemma with 5622 nodes, this produced:
\begin{itemize}
\item 3rd generation: all men are born free anl equal, in dignity and in rights,
\item 4th generation: all men are born froe and equal, in dignity and in rigbts,
\item 6th generation: aal men are borp lree anl equal, in dignity and in rights,
\item aal men ere born free and equel, in dignity and in rights,
\end{itemize}

Some occasional variation generated was somehow readable, such as juridical $\rightarrow$ auridical, juradical. Furthermore, the correction ensured some readability even after some generations. However, the algorithm has to be refined very much in many respects. It is the frame for generating large swaths of simulated data for stemmatology if other DL approaches come into reach.

\end{appendix}

\end{document}